%% file: main.tex
\documentclass[runningheads]{llncs}
\usepackage[T1]{fontenc}
\usepackage{graphicx}
\usepackage[caption=false]{subfig}
\usepackage{pdfpages}
\usepackage{framed}
\usepackage{xcolor}
\usepackage{cite}
\usepackage{hyperref}
\usepackage{xurl}
\usepackage[english]{babel}
\usepackage[misc,geometry]{ifsym}
\usepackage[shortcuts,acronym]{glossaries}
\usepackage{booktabs}
\usepackage{tabularx}

\begin{document}
\title{Task Planning Support for Arborists and Foresters: Comparing Deep Learning Approaches for Tree Inventory and Tree Vitality Assessment Based on UAV-Data\thanks{Special thanks to our cooperation partner Smart City Bamberg. The project BaKIM is supported by Kommunal?Digital! funding of the Bavarian Ministry for Digital Affairs. Project funding period: 01.01.2022 - 31.03.2024.}}%

\titlerunning{Task Planning Support for Arborists and Foresters}%
% If the paper title is too long for the running head, you can set
% an abbreviated paper title here
%
\author{Jonas Troles \Letter \inst{1}\orcidID{0000-0001-8686-5168} \and
Richard Nieding\inst{1} \and \\
Sonia Simons\inst{2} \and
Ute Schmid\inst{1}\orcidID{0000-0002-1301-0326}}%
\authorrunning{J. Troles et al.}%
% First names are abbreviated in the running head.
% If there are more than two authors, 'et al.' is used.
%
\institute{University of Bamberg, Cognitive Systems Group, 96049 Bamberg, Germany
\email{jonas.troles@uni-bamberg.de} \and
TU Berlin, Straße des 17. Juni 135, 10623 Berlin, Germany}%
\maketitle% typeset the header of the contribution
\begin{abstract}%
Climate crisis and correlating prolonged, more intense periods of drought threaten tree health in cities and forests. In consequence, arborists and foresters suffer from increasing workloads and, in the best case, a consistent but often declining workforce. To optimise workflows and increase productivity, we propose a novel open-source end-to-end approach that generates helpful information and improves task planning of those who care for trees in and around cities. Our approach is based on RGB and multispectral UAV data, which is used to create tree inventories of city parks and forests and to deduce tree vitality assessments through statistical indices and Deep Learning. Due to EU restrictions regarding flying drones in urban areas, we will also use multispectral satellite data and fifteen soil moisture sensors to extend our tree vitality-related basis of data. Furthermore, Bamberg already has a georeferenced tree cadastre of around 15,000 solitary trees in the city area, which is also used to generate helpful information. All mentioned data is then joined and visualised in an interactive web application allowing arborists and foresters to generate individual and flexible evaluations, thereby improving daily task planning.

\end{abstract}%
\begingroup
\let\clearpage\relax
\include{Content/1-Introduction}
\include{Content/2-Data-Acquisition}
\include{Content/3-0-Tree-Inventory}
\include{Content/3-1-Tree-Inventory}
\include{Content/3-2-Tree-Inventory}
\include{Content/4-Tree-Health-Estimation}

\include{Content/5-Interactive-Web-Application}
\include{Content/6-Conclusion}
\endgroup
%\include{Content/Springer-LNCS-Graveyard}
%\include{bibliography}
%
% ---- Bibliography ----
%
% BibTeX users should specify bibliography style 'splncs04'.
% References will then be sorted and formatted in the correct style.
%
\bibliographystyle{splncs04}
\bibliography{bibliography-bibtex}

\end{document}

%% file: Content/1-Introduction.tex
\section{Introduction}%

\begin{figure}%
    \centering
    \centering
        \includegraphics[width=\textwidth]{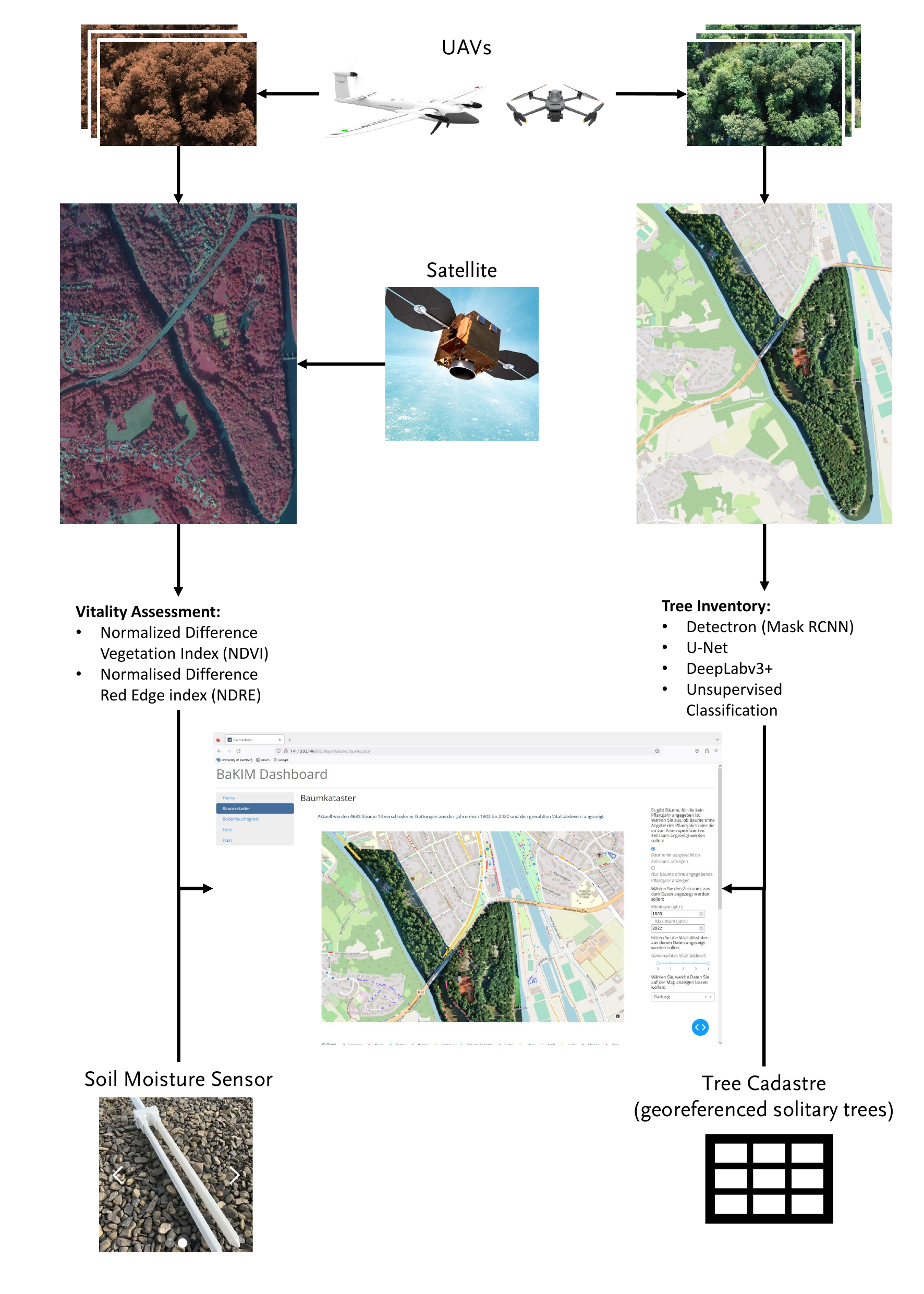}%
    \caption{Overview of the BaKIM pipeline with images from~\cite{trinity, dji-m3m, pleiades, agvolution}}%
    \label{fig:BaKIMOverallPipeline}%
\end{figure}%

According to the IPCC report from 2022~\cite{ipcc2022}, the climate crisis is an enormous challenge and already affects the lives of billions of people. One of the many negative impacts of the climate crisis is its threat to whole ecosystems and, therefore, the health of trees and forests by rising temperatures and consecutive drought events, as experienced from 2018 to 2020 and in 2022 in Germany~\cite{hess-drought, afz-waldzustand, kehr-Folgeschaeden}. The consequence is an increasing vulnerability to a variety of pests and increasing tree mortality rates, leading to an increasing frequency and scope of measures arborists and foresters have to take~\cite{Raum-Auswirkungen}. At least in Bamberg, but presumably in most German cities, this has to be done without an increase in staff in the teams of arborists and foresters because the funding for said departments is seldom increased. On top, they suffer from a shortage of skilled workers.

To tackle these problems, in 2021, a cooperation of \emph{Smart City Bamberg}, Bamberg's lead arborist, Bamberg's lead forester and the \emph{Chair of Cognitive Systems} at the \emph{University of Bamberg} applied for project funding to support Bamberg's arborists and foresters with Deep Learning approaches based on UAV-data (Unmanned Aerial Vehicle). The result is the project \emph{BaKIM}, which aims to generate helpful and flexible information from remote sensing data gathered by UAVs and satellites, as well as data from soil moisture sensors. To generate information from this data, an ensemble of Deep Learning, Machine Learning and statistical methods is used. All information is then visualised in an interactive web application which offers basic Geographic Information System (GIS) functionalities and extends these by easy and fast filtering and plotting. The complete pipeline of BaKIM is illustrated in Fig.~\ref{fig:BaKIMOverallPipeline}.

In addition to these goals, it is of great importance for our project to develop a human-centred AI approach as reflected in~\cite{waefler2020explainability}. BaKIM will not, and does not try to be an autonomous AI approach which replaces the decision-making of arborists and foresters. Quite the contrary, BaKIM will rely on arborists and foresters for: (i) verifying its AI-based decision suggestions, (ii) improving BaKIM, (iii) learning from BaKIM, and (iv) taking responsibility for the final decision as well as for compliance with legislation and ethical standards~\cite{waefler2020explainability, samek2017explainable}.

%% file: Content/2-Data-Acquisition.tex
\section{Data Acquisition and Technology}%
The first and most important data source for BaKIM are two drones which produce high-resolution image data. As EU-regulation~\cite{eu-commission} forbids the use of UAVs heavier than 250g over houses and uninvolved people, we decided to additionally use satellite imagery for the health assessment of solitary city trees. To control the assessment of drought stress in trees, 15 soil moisture sensors were installed in the city. In the following, the different data acquisition tools are described.

\subsection{Areas of Interest}%
\begin{figure}%
    \centering
    \includegraphics[width=\textwidth]{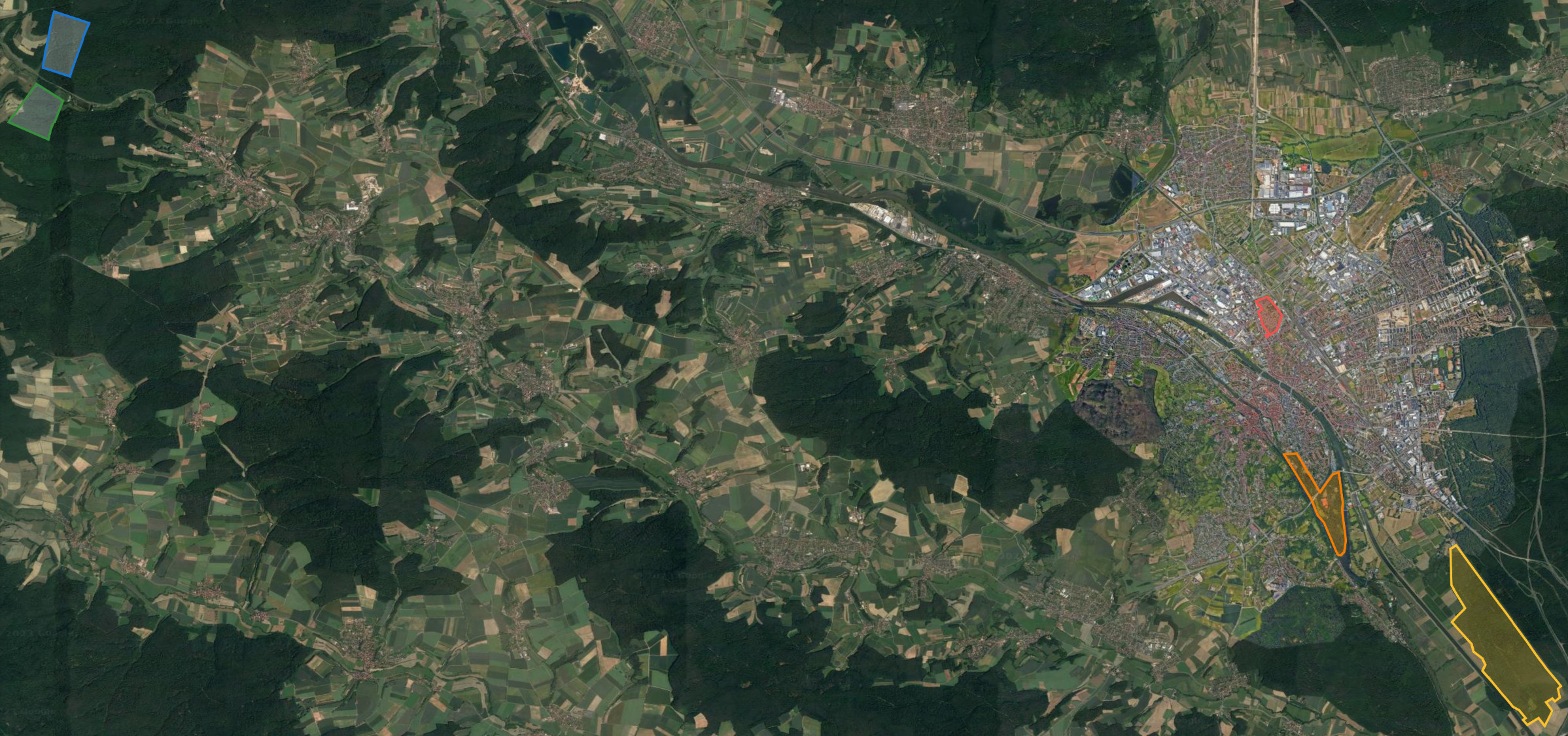}%
    \caption{All AOIs in and around Bamberg (Top left to bottom right: Tretzendorf AOI 1 (blue), Tretzendorf AOI 2 (green), Bamberg Graveyard AOI (red), Bamberg Hain AOI (orange), Bamberg Stadtwald AOI (yellow))}%
    \label{fig:bakim-aois}%
\end{figure}

In BaKIM, we defined several Areas of Interest (AOIs), which we monitor throughout the project and partially sample ground truth labels on a single tree basis to train our different Convolutional Neural Nets (CNNs). See Fig.~\ref{fig:bakim-aois} to get an overview of the size and location of the following AOIs:
\newpage

% TODO: Check Umbruch
\begin{itemize}
    \item Forest AOIs:
    \begin{itemize}
        \item[-] Stadtwald AOI: 190ha, mostly coniferous forest
        \item[-] Tretzendorf AOI 1: 60ha, mixed forest
        \item[-] Tretzendorf AOI 2: 45ha, mixed forest
    \end{itemize}
    \item Bamberg Hain AOI: 50ha, park with deciduous forest-like areas
    \item Bamberg graveyard AOI: 15ha solitary trees
    \item Bamberg city AOI: 5,462ha mostly solitary trees (only satellite imagery)
\end{itemize}

% UAV DATA
\subsection{UAV-Data}%
\begin{figure}%
    \centering
    \subfloat[\centering Trinity F90+~\cite{trinity}]{{\includegraphics[width=0.46\textwidth]{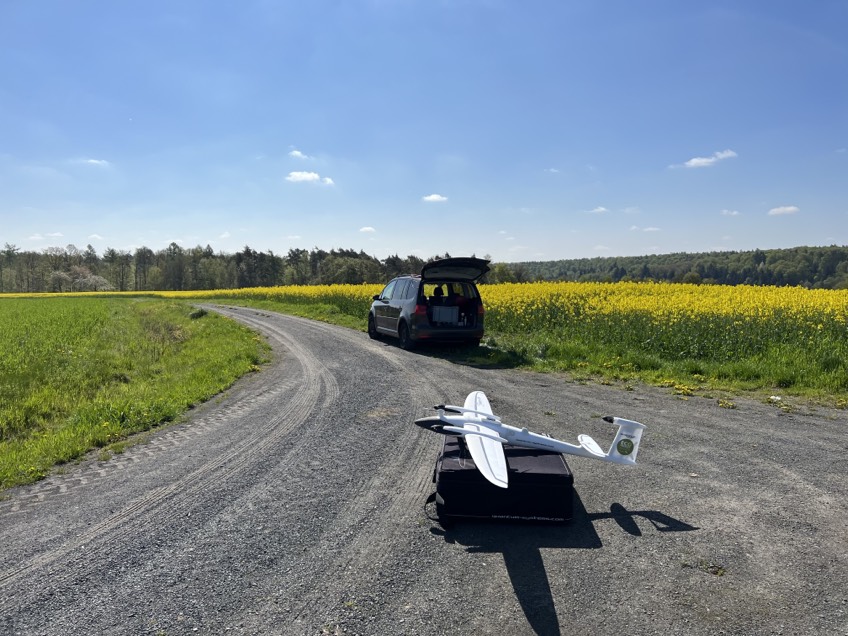}}}%
    \qquad
   \subfloat[\centering DJI M3M~\cite{dji-m3m}]{{\includegraphics[width=0.46\textwidth]{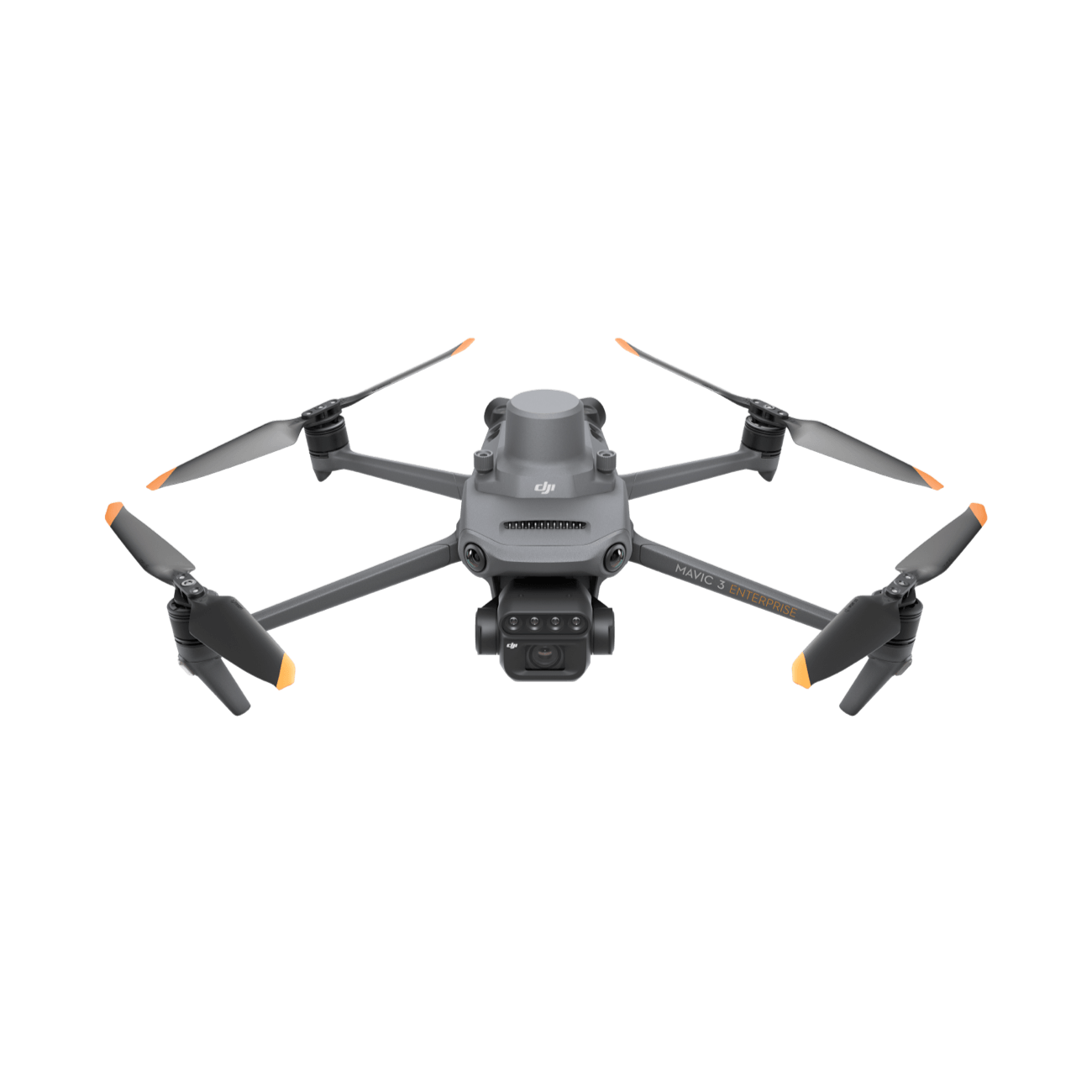} }}%
    \caption{Both UAVs used for data acquisition in the BaKIM project}%
    \label{fig:Drones}%
\end{figure}

Due to the mentioned EU-regulation, we use two different UAVs: One fixed-wing UAV with a take-off weight of up to 5,500g, which we use for forest areas outside of the city and one smaller quadrocopter with a take-off weight of up to 1,050g which we use for the Hain. Both UAVs are shown in Fig.~\ref{fig:Drones}.
\vspace{10pt}

% TRINITY F90+
\noindent \textbf{Trinity F90+:} 
The Trinity F90+ is a fixed-wing UAV with Vertical Take-Off and Landing (VTOL) capability and a flight time of about 90 minutes. With its 5,500g take-off weight and 239cm wingspan, it is a C3 UAV and must be flown in the open A3 category, meaning it must not fly near people and must fly outside urban areas (150 m distance). However, it can carry interchangeable payloads with heavier, more advanced sensors and cover larger areas than smaller UAVs. Therefore, we use it for the three more remote forest AOIs outside of Bamberg.

The first sensor, used for RGB data collection, is a Sony RX1-RII camera with 42.4 MP and a theoretical Ground Sampling Distance (GSD) of 1.55cm when flying 120m Above Ground Level (AGL). Our experience shows we get a GSD of about 1.6 - 1.9cm in the finished orthomosaic.
The second sensor, used for multispectral (MS) data collection, is a MicaSense Altum-PT with 3.2MP per MS band and a thermal infrared sensor with a resolution of 320 x 256 pixels. The resulting GSD is 5.28cm for the MS bands and 33.5cm for the thermal band when flying 120m AGL. Due to supply difficulties, our MicaSense Altum-PT sensor arrived later than expected and was not tested yet.
\vspace{10pt}

% DJI MAVIC 3 MULTISPECTRAL
\noindent \textbf{DJI Mavic 3 Multispectral:}
To cover RGB and MS data acquisition of the Hain, we use the smaller and lighter quadrocopter DJI Mavic 3 Multispectral (DJI M3M), released in Q1 of 2023. Compared to the Trinity F90+, it does not support swappable payloads and has a much shorter flight time of 43 minutes at max. In return, with its maximum take-off weight of 1,050g and a diagonal length of 38cm, it is a C2 UAV and can be flown in the A2 category where a flight in urban areas and near uninvolved people is possible.
The built-in sensor consists of a 20MP RGB sensor with 2.95cm GSD at 120m AGL and a 5MP multispectral sensor with 5.07cm GSD at 120m AGL. 
\vspace{10pt}

% FORTRESS
\noindent \textbf{FORTRESS Dataset:}
To start the training of the different Deep Learning approaches described in Sect.~\ref{sec:DeepLearning} as early as possible, we used the FORTRESS dataset~\cite{schiefer-fortress}. It consists of 47ha of very-high-resolution orthomosaics with a GSD of up to 0.6cm and covers mostly coniferous forest in the southwest of Germany. In total, 16 different classes are labelled in FORTRESS. For a more detailed description of the dataset, see the paper of Schiefer et al.~\cite{schiefer2020mapping}.

% ORTHOMOSAIC GENERATION
\subsection{Orthomosaic Generation}
To get image data where every pixel is georeferenced and the perspective is corrected to a nadir view, the single UAV images need to be processed. For this, different software is available, and we tested two products: \emph{WebODM}\footnote{\url{https://github.com/OpenDroneMap/WebODM}} which is open source and free software as well as \emph{Agisoft Metashape}\footnote{\url{https://www.agisoft.com/}} which is commercial software. After testing and comparing WebODM and Metashape, we found the orthomosaics produced with Metashape to be of slightly higher quality, showing fewer artefacts. This was especially the case for our images taken with the Trinity F90+, as they lack the very high front overlap necessary for orthomosaic generation. You can find examples of Hain orthomosaics in Figs.~\ref{fig:BaKIMOverallPipeline} and~\ref{fig:tree-cadastre}.
\vspace{10pt}

% IMAGE OVERLAP
\noindent \textbf{Image Overlap:} Image overlap is the most crucial parameter for orthomosaic generation. Especially when it comes to forests with their small structures (leaves, branches) and the abrupt height changes of the surface (trees, ground). Therefore, a front overlap of 90-95\% and a sidelap of 80\% is recommended. While the DJI M3M can change its flight speed, the Trinity F90+ must keep an airspeed of 17m/s to stay airborne. As the maximum shutter speed of the sensors is limited, this results in a maximum front overlap of about 70\% for the Sony RX1-RII payload of the Trinity F90+. We partially compensate for this through a sidelap of 90\%, but the resulting orthomosaics of the Trinity F90+ still show artefacts in certain points of the forest areas.

% GROUND TRUTH LABELS
\subsection{Ground Truth Labelling}\label{sec:ground-truth-labelling}
Especially for the individual tree crown delineation (ITCD) described in Sect.~\ref{sec:singletreedetection}, but also for tree species prediction, we need ground truth data on the tree instance level. Therefore, we commissioned a forester to delineate 108ha of tree crowns in the AOIs Stadtwald, Tretzendorf and Hain. Additionally to the delineation, tree species and a rough vitality assessment is labelled.

\subsection{Other Data}
% SATELLITE DATA
\noindent \textbf{Satellite Data:}
As already mentioned, EU-law heavily restricts flying UAVs in urban areas. Therefore, we will use multispectral satellite imagery taken by the \emph{Airbus pléiades neo} satellite with a GSD of 1.2m. This imagery is tasked for July 2023 and September 2023 and will be used for additional tree health assessment.
\vspace{10pt}

% SOIL MOISTURE SENSORS
\noindent \textbf{Soil Moisture Sensors:}
To gather ground truth data on drought stress, we decided to use soil moisture sensors provided by \emph{Agvolution}. They track the soil's water content in three different depths and send the data via \emph{mioty} standard to agvolution's server, where they are then processed and made accessible via an API.
\vspace{10pt}

\begin{figure}%
    \centering
    \includegraphics[width=\textwidth]{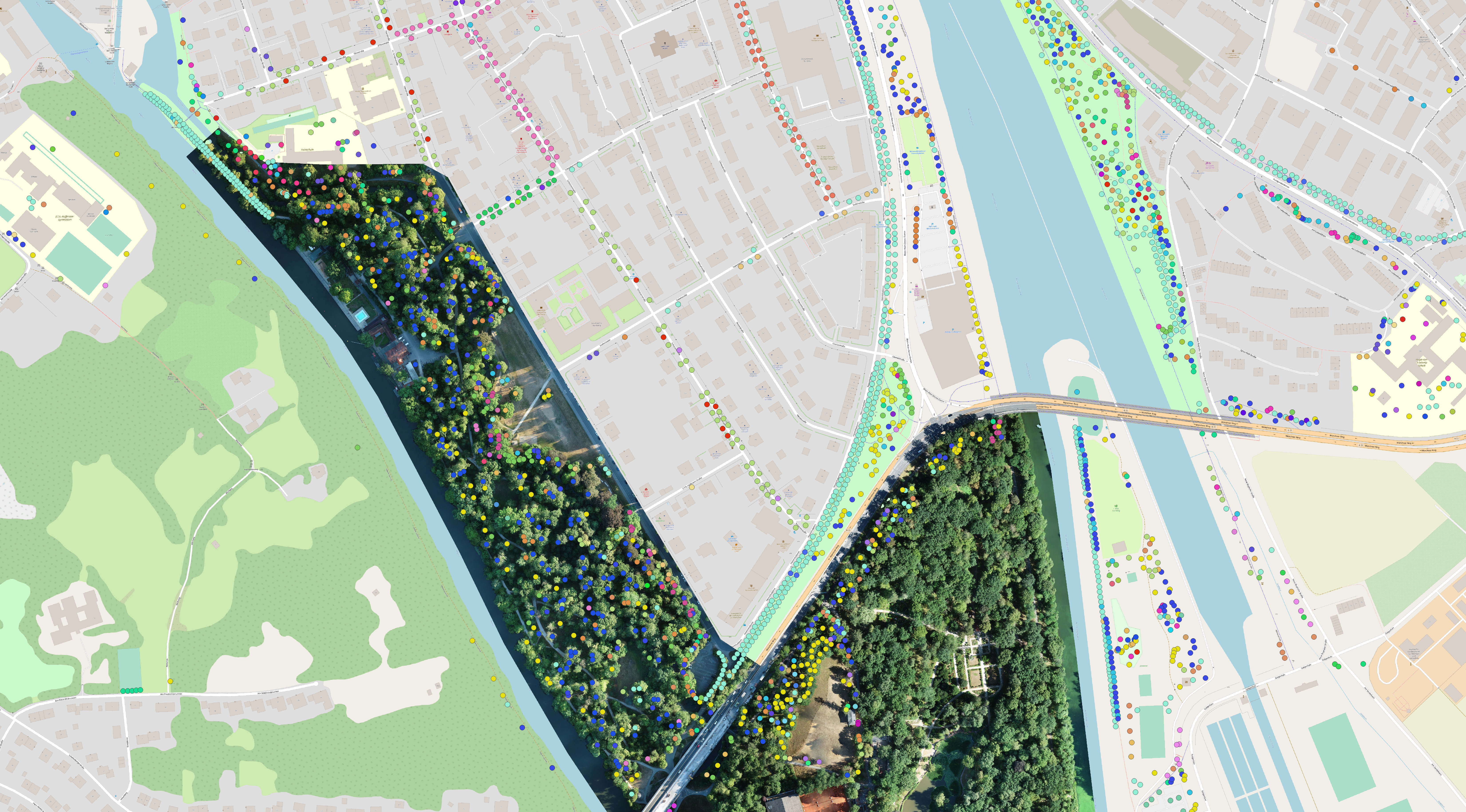}%
    \caption{One possible visualisation of tree cadastre data with the point colour depicting the tree species}%
    \label{fig:tree-cadastre}%
\end{figure}

% TREE CADASTRE
\noindent \textbf{Tree Cadastre:}
Bamberg's arborists have tracked the vitality of most of the solitary trees in the city area for decades and visit every tree at least once a year. This information is stored in the \emph{tree cadastre} and contains, among other things, georeferences, tree species, tree dimension estimations, tree vitality and site information. Unfortunately, the used software does not save a history of this information per tree location, and no historical backups are available. Therefore, we started to back up the current version every six months to make future time series analysis possible. Figure~\ref{fig:tree-cadastre} shows a part of the Hain orthomosaic and tree cadastre information in the form of coloured dots.

%% file: Content/3-0-Tree-Inventory.tex
\section{Methods of Tree Inventory Generation}\label{sec:DeepLearning}%
While solitary trees in the city of Bamberg are all georeferenced and tracked well, this does not apply to the Hain and the forest areas surrounding Bamberg. The latter are inventoried every 10-20 years by an estimation procedure, and the Hain is partially inventoried because trees near paths are checked regularly. Therefore, according to our forester and arborist, a more frequent and more accurate park and forest inventory procedure supports task planning and allows for a faster reaction to diseases like bark beetle infestation or weather events like storms. To accomplish this, we decided to use high-resolution orthomosaics with a Ground Sampling Distance (GSD) of 1.5 to 2.0 cm per pixel and different Deep Learning and Machine Learning approaches from the Computer Vision domain, which are described in the following sections.

%% file: Content/3-1-Tree-Inventory.tex
\subsection{Single Tree Detection}\label{sec:singletreedetection}%

Single tree detection is divided into two categories: Individual Tree Detection (ITD), which is called \emph{object detection} in the domain of computer vision (CV) and individual tree crown delineation (ITCD), which is called \emph{instance segmentation} in the domain of CV. ITCD is considered the more challenging task, as it additionally determines the crown boundaries of individual trees.

Historically, tree detection was accomplished with unsupervised approaches such as local maxima (LM)~\cite{monnet2010, Pitkaenen2001}, marker-controlled watershed segmentation (MCWS)~\cite{Liao2022}, region-growing~\cite{Erikson2005}, edge detection~\cite{wang2004individual} and template matching~\cite{Huo2020, Larsen2011}, with LM and MCWS methods being the most commonly used~\cite{Yu2022}. The rapid development of deep learning methods, especially convolutional neural networks (CNNs), led to much better results in CV tasks due to their ability to extract low and high-level features. This motivated the application of deep learning methods in remote sensing applications. For example, the object detection method Faster R-CNN~\cite{Weinstein2020a} is used for ITD tasks, and more recently, the instance segmentation method Mask R-CNN~\cite{Ball2022, Hao2021, Lucena2022, Yang2022} is used for ITCD tasks. The difference between these approaches is that Faster R-CNN draws a bounding box around the detected single tree. Mask R-CNN uses the Faster R-CNN structure and extends the bounding box prediction by a branch to predict a segmentation mask. This is advantageous because it provides more accurate information about the actual crown area and avoids or greatly reduces distortions that may occur in the background area of the bounding box.

Both classical and more recent deep learning methods have advantages and disadvantages in processing time and data requirements. The most commonly used classical methods, such as LM and MCWS~\cite{Yu2022}, require a distinct height model to perform their methods. Faster R-CNN and Mask R-CNN, on the other hand, need RGB image data to detect trees. Another disadvantage of classical methods is that they show difficulties in detecting trees in areas where tree crowns overlap strongly~\cite{Yu2022}. In contrast, Yu~\cite{Yu2022} has shown that LM and MCWS are less computationally intensive than Mask R-CNN. Another advantage of these classical methods is that they do not require manually labelled data and, therefore, no training phase to detect trees. Faster R-CNN and Mask R-CNN rely on ground truth data and mostly some form of retraining to perform the detection task on new datasets. This makes it clear that the tree detection method should be chosen according to the corresponding forest structure and in accordance with the desired goal.

Due to the higher accuracies of DL approaches in BaKIM, a retrained tree detection model based on the Mask R-CNN \emph{detectree2} implementation is used~\cite{Ball2022}. Figure~\ref{fig::pipeline} shows predictions for a section of the FORTRESS dataset.

\subsection{Unsupervised Tree Species Classification}%
\begin{figure}
    %\begin{framed}
	\includegraphics[width=1\textwidth]
        {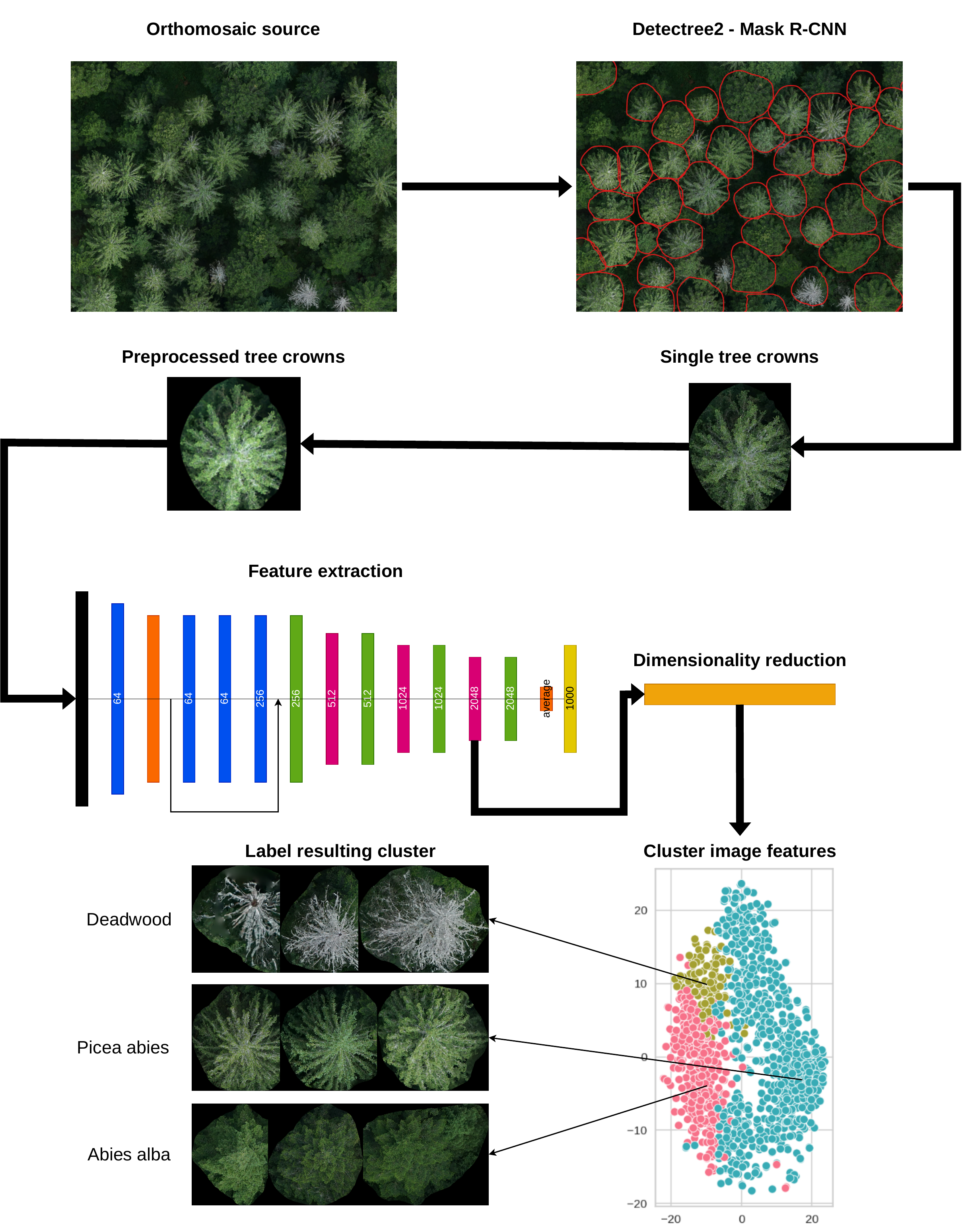}
     %\end{framed}
        \caption{The pipeline used for the unsupervised classification of single tree crown images}
	\label{fig::pipeline}
\end{figure}
Many UAVs are characterised by relatively low cost and a simple structure. Consequently, the dissemination and usage of UAVs increases, and much high-resolution image data is captured~\cite{Hanapi2019, Yang2022}. This makes the demand for technologies and solutions that can quickly process this data more urgent. Most, if not all, current approaches that classify tree species rely heavily on a domain expert to determine tree species for every single tree of the training data in a supervised manner. This cumbersome human labelling process may not keep up with the increasing demand for solutions.

The inclusion of an unsupervised method for classifying tree species is rarely an issue in the literature. Franklin used multispectral data and pixel-based and object-based image analysis to classify tree species in an unsupervised way~\cite{Franklin2018}. Another unsupervised method proposed by Gini et al. also works pixel-based~\cite{Gini2014}. Schäfer et al. used spectroscopy data to identify tree species via clustering without knowing how many trees per species occur~\cite{Schaefer2016}. All these unsupervised methods are based on costly spectral or LiDAR data that may not be available to every practitioner. They are also unable to map a species to a particular detected tree.

It is promising to use an unsupervised classification approach via clustering to overcome the drawbacks mentioned above. The presented pipeline is inspired by the work of~\cite{Cohn2021, Gradisar2023, Ji2022} with the goal of transferring their good classification results to the field of single tree classification. The first step in this pipeline is to determine single tree crowns in an orthomosaic, as shown in Fig.~\ref{fig::pipeline}. Subsequently, small single tree crown images are extracted from the original orthomosaic. Then the preprocessing techniques are applied to these small images. Next, the image features are extracted from these small images. A dimensionality reduction step follows to reduce the features to more meaningful components. Lastly, a clustering algorithm is used, which assigns a corresponding label to each tree image. In this setup, the domain expert only needs to classify the appearing tree classes once, which can be done quite quickly compared to determining the species of every single tree in a particular training set.

A retrained tree detection model based on the Mask R-CNN \emph{detectree2} implementation of~\cite{Ball2022} was used to detect single tree crowns at the beginning of this pipeline. This model was retrained on a total of 4539 manually delineated tree crowns from orthomosaics: 2262 from the FORTRESS dataset~\cite{schiefer2020mapping}, 1651 from the Bamberg Stadtwald AOI, and 626 from the Tretzendorf AOI 1.

\begin{figure}
	\includegraphics[width=1\textwidth]{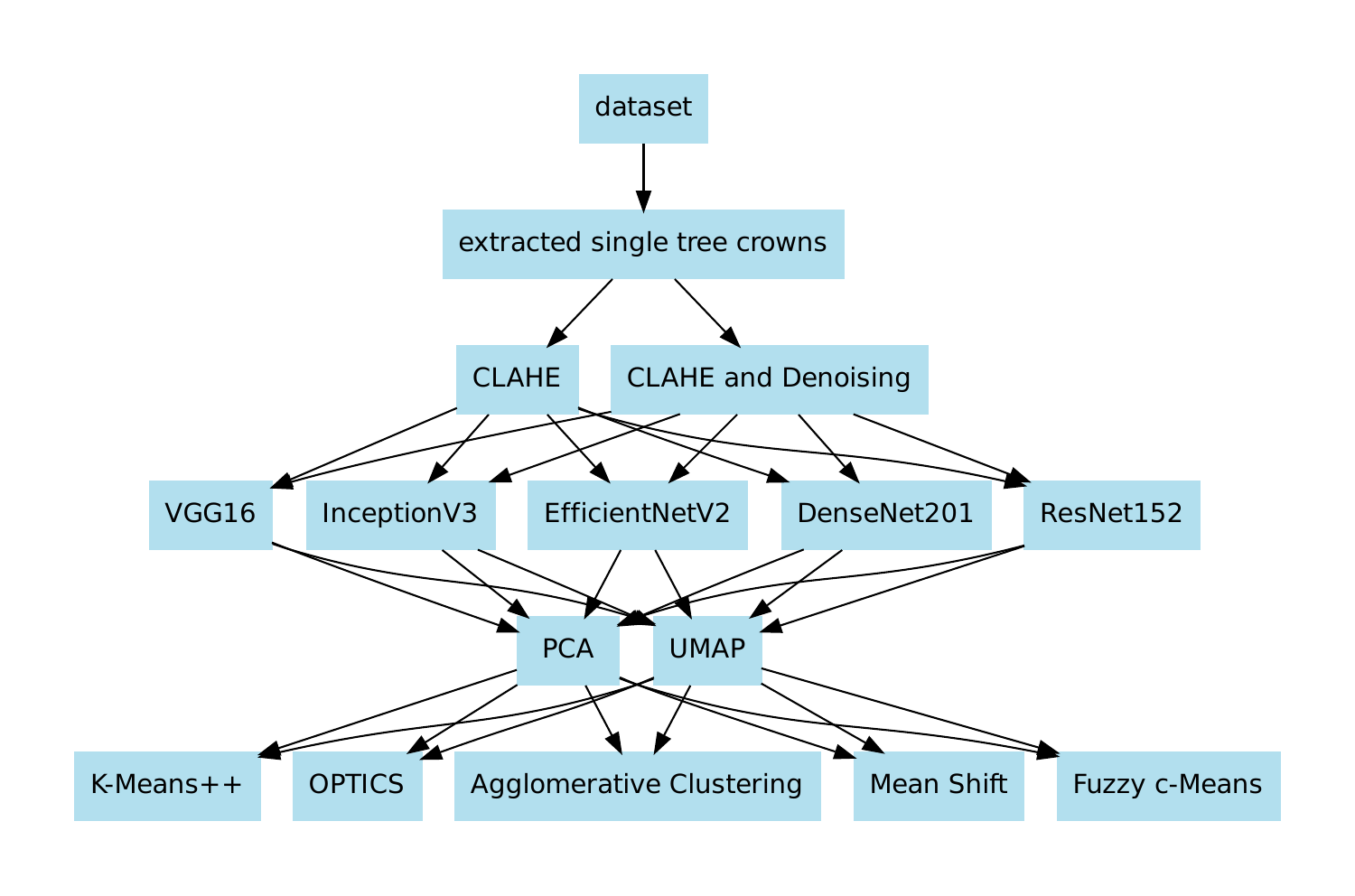}
	\caption{Scheme representing all possible combinations of methods}
	\label{fig::combinations}
\end{figure}

This pipeline's preprocessing, dimensionality reduction, and clustering steps can be performed in several ways, as shown in Fig.~\ref{fig::combinations}. Therefore, an experiment is conducted to determine and evaluate the best combination of methods. Two different preprocessing steps were applied to the image datasets in this experiment. Once Contrast Limited Adaptive Histogram Equalization (CLAHE) was applied to the images, and once a combination of CLAHE and denoising was applied to the images. Afterwards, the image features were extracted via a forward hook using the following pretrained CNN backbones: VGG16, ResNet152, InceptionV3, EfficientNetV2 and DenseNet201. PCA and UMAP were used for the dimensionality reduction of the image features. The reduced feature vectors were used as input for the cluster algorithms k-means++, mean shift, fuzzy c-means, agglomerative clustering and Ordering Points to Identify the Clustering Structure (OPTICS). Some cluster algorithms such as k-means++, fuzzy c-means (fc-means) and agglomerative clustering require setting the number of clusters in advance, which is difficult. However, in this case, the number of clusters was set to the number of occurring species to obtain more meaningful results. All mentioned methods for each step result in 100 possible combinations.

Evaluating cluster results is challenging in real-world applications because ground truth labels are often unavailable. In this case, however, ground truth data was available, and therefore cluster algorithms could be evaluated based on the F1-scores. The proposed experiment was performed 30 times, and then the mean F1-scores were calculated from these 30 cumulative values for each possible combination of methods.

\begin{table}
	\caption{Five best method combinations from the experiment conducted on the FORTRESS dataset}
	\centering
        \setlength{\tabcolsep}{4pt}
	\label{table::results}
	\begin{tabular}{lcccccc}
		\toprule
		  Preprocess &       CNN &   DR & Clustering &  Species &  F1 & weighted F1 \\
		\midrule
		clahe &  densenet &  pca &  k-means++ & 4 & 0.82 & 0.79 \\
		clahe+denoising &  densenet &  pca &  k-means++ & 4 & 0.79 & 0.75 \\
		  clahe &    resnet & umap &   fc-means & 4 & 0.75 & 0.72 \\
		clahe & resnet & umap &  k-means++ & 4 & 0.75 & 0.72 \\
		clahe &    resnet & umap &      agglo & 4 & 0.75 & 0.72 \\
		\bottomrule
	\end{tabular}
\end{table}

\begin{table}
    \caption{F1-scores per class for the best combination of methods, clustering the data in four classes}
    \centering
    \setlength{\tabcolsep}{5pt}
    \begin{tabular}{lcccc}
        \toprule
        Class                 & TP   & FP  & FN  & F1   \\
        \midrule
        Picea Abies           & 1534 & 264 & 123 & 0.89 \\
        Fagus Sylvatica       & 589  & 192 & 89  & 0.81 \\
        Pinus Sylvestris      & 165  & 30  & 64  & 0.78 \\
        Abies Alba            & 253  & 70  & 87  & 0.76 \\
        Pseudotsuga Menziesii & 0    & 0   & 113 & 0.00    \\
        Larix Decidua         & 0    & 0   & 30  & 0.00    \\
        Quercus Spec          & 0    & 0   & 23  & 0.00   \\
        deadwood              & 0    & 0   & 15  & 0.00    \\
        Fraxinus Excelsior    & 0    & 0   & 9   & 0.00    \\
        Betula Pendula        & 0    & 0   & 3   & 0.00    \\
        \bottomrule
    \end{tabular}
    \label{tab:f1-score-best}
\end{table}

The results of the best-performing combination of methods were evaluated using FORTRESS as an example.
The five most frequent tree species after the extraction of 3097 single tree crowns were Picea Abies (53.5\%), Fagus Sylvatica (21.9\%), Abies Alba (11.0\%), Pinus Sylvestris (7.4\%) and Pseudotsuga Menziesii (3.6\%). Table~\ref{table::results} shows that the best combination is composed of CLAHE, densenet, PCA, and k-means++. The F1-score and weighted F1-score show only slight differences, even though only four of ten classes were clustered. Nevertheless, the large amount of correctly predicted samples from the majority classes has a higher impact on the F1-score than those with a lower sample size. These findings suggest that the proposed pipeline has problems in classifying minority classes. Table~\ref{tab:f1-score-best} shows how only four classes are assigned, which yields high F1-scores for the majority classes, but an F1-score of 0.00 for all unassigned classes.

This unsupervised pipeline demonstrates that tree species can be classified by clustering based on RGB image data alone. However, the inability to detect lower sample classes reduces the practical benefit if a practitioner wants a correct picture of the species distribution. Moreover, it became clear that it is difficult to clearly distinguish the tree crowns of different species because of their similar structure. The results also show that the preprocessing steps are not sufficiently capable of highlighting the particular characteristics of the tree species and that the separation of the clusters needs to be improved in the future.

%% file: Content/3-2-Tree-Inventory.tex
\subsection{Semantic Segmentation of Tree Species}%
Another way to classify tree species in high-resolution UAV imagery is semantic segmentation. It resembles an ITCD approach that does not distinguish between individual trees but uses less computational resources than, for example, the Mask R-CNN based \emph{detectree2}~\cite{schiefer2020mapping, Ball2022}.

In~\cite{schiefer2020mapping}, Schiefer et al. explored the viability of using convolutional neural networks (CNNs) for classifying tree species from drone images. They used an adjusted version of U-Net~\cite{ronneberger2015u}, which was originally designed for the segmentation of biomedical images but has found its way into other areas like semantic segmentation of tree species~\cite{kattenborn2021review}.
It is made up of two symmetric paths, the encoder and the decoder. The encoder downsamples the input image to capture contextual information, while the decoder upsamples the encoded feature maps to increase the spatial resolution. One key feature of U-Net is that the decoder also receives skip connections from the encoder. This way, the information learned by the downsampling layers in the encoder is used to reconstruct the input image at its original spatial resolution while preserving fine details and features.

The FORTRESS dataset~\cite{schiefer-fortress} includes 16 classes. After analysing the share of pixels each class has in the entire dataset, we noticed that the smallest seven classes (\textit{other, Aesculus, Fallopia, Ilex, Fraxinus Excelsior, Larix Decidua} and \textit{Betula Pendula}) combined only makeup 1.3\% of the total area covered, while the largest class alone (\textit{Picea Abies}) makes up 38.3\%.
This substantial imbalance in the dataset was already mentioned in~\cite{schiefer2020mapping}, where they tried to mitigate this by using weighted cross-entropy loss. On top of a weighted loss function, we also decided to combine the smallest seven classes into just one class called \textit{other} to adapt to this extreme imbalance in the dataset.

The primary metric we used to evaluate the performance of the models is Intersection over Union (IoU). 
It measures the accuracy of a prediction as a value between 0 and 1, where 0 means no overlap between the prediction and the ground truth and 1 being an exact match. 
The overall IoU is calculated as the average of the IoU of each class, weighted by their share in the dataset. The overall IoU reaches around 0.78 after 100 epochs of training. Despite achieving a good overall result, examining the per-class IoU values shows a significant difference in the model's performance across classes that are less represented in the dataset than those that are well-represented.
Table~\ref{tab:per_class_iou} shows that classes \emph{Picea Abies}, \emph{Fagus Sylvatica} and \emph{Abies Alba} make up 75.5\% of the dataset and have IoU Test scores of 0.75, 0.73 and 0.76 respectively. While classes \emph{Quercus Spec.} and \emph{Acer Pseudoplatanus} make up 1.9\% and have much lower IoU Test scores of 0.11 and 0.39. However, these low representations of some classes will be reduced with the labelled data from the Bamberg AOIs.

\begin{table*}
    \caption{Per class comparison of area-related share and IoU of the validation-set and test-set}
    \centering
    \label{tab:per_class_iou}
    \setlength{\tabcolsep}{4pt}
    \begin{tabular}{lcccc}
        \toprule
        class                 & area-related share    & IoU Validation & IoU Test  &  \\
        \midrule
        Picea Abies                & 38.3\%              & 0.84 & 0.75 &  \\
        Fagus Sylvatica            & 26.5\%              & 0.82 & 0.73 &  \\
        Forest floor               & 12.6\%              & 0.65 & 0.54 &  \\
        Abies Alba                 & 10.7\%              & 0.76 & 0.61 &  \\
        Pinus Sylvestris           & 4.2\%               & 0.78 & 0.71 &  \\
        Pseudotsuga Menziesii      & 3.5\%               & 0.77 & 0.81 &  \\
        Acer Pseudoplatanus        & 1.1\%               & 0.46 & 0.39 &  \\
        Quercus spec.              & 0.8\%               & 0.32 & 0.11 &  \\
        Deadwood                   & 0.7\%               & 0.30 & 0.18 &  \\
        other                      & 1.3\%               & 0.27 & 0.32 &  \\
        \midrule
        weighted average           &                     & 0.78 & 0.68 & \\
        %overall                    &                       & xx     & xx     &  \\
        %average                    &                       &        &        & \\
        \bottomrule
    \end{tabular}
\end{table*}

As part of future work, we aim to incorporate a post-processing step that was employed in the accuracy assessment of~\cite{schiefer2020mapping}. This involves generating multiple predictions for each pixel using a moving window technique and assigning the final prediction through a majority vote.

We are also currently working on using another CNN for the same task, Deeplabv3+~\cite{chen2018encoder}, which is a promising CNN architecture for this task as shown by~\cite{morales2018automatic, ferreira2020individual, lobo2020applying}. 
Deeplabv3+ is based on Deeplabv3~\cite{chen2017rethinking}, which uses atrous convolutions~\cite{holschneider1990real} and atrous spatial pyramid pooling (ASPP)~\cite{chen2017deeplab}. Atrous convolutions are used to retain the same spatial resolution while increasing the feature maps without increasing the parameters or amount of computation. ASPP aggregates features at multiple scales and captures context information from a larger image region.

Deeplabv3+ combines Deeplabv3 with an Encoder-Decoder structure like U-Net. It uses a slightly modified version of Deeplabv3 as the encoder. Instead of directly upsampling the encoder output to the original image resolution, the decoder combines them with features from early convolutional layers, analogous to the skip connections used in U-Net. While Deeplabv3+ will most likely outperform U-Net, it also needs more computational resources.

%% file: Content/4-Tree-Health-Estimation.tex
\section{Tree Vitality Assessment}%
Apart from detecting single trees and their species, assessing tree vitality on an instance level is of great importance to arborists and foresters. To accomplish this, we opted for two different kinds of approaches: First, statistical evaluation of multispectral sensor imagery by building indices and second, deep learning methods applied to very-high-resolution imagery to detect dead branches or secondary pests like mistletoes.

\subsection{Statistical Indices Derived from Multispectral Data}%
The usage of multispectral imagery for tree vitality assessment dates back to the 1970s and has been used since~\cite{rouse1974monitoring, senoo1988assessment, moran1994estimating, ismail2007forest, lausch2013forecasting, huo2023assessing}. In BaKIM, we plan to use at least the following two indices:
\vspace{10pt}

\noindent \textbf{NDVI:} The Normalised Difference Vegetation Index was introduced back in 1974 and is one the most used indices to classify if a pixel shows live green vegetation~\cite{rouse1974monitoring}. It is based on the near-infrared (NIR) band as well as the red band and is calculated as follows:
\begin{equation}
    \mathrm{NDVI}=\frac{(\mathrm{NIR}-\mathrm{Red})}{(\mathrm{NIR}+\mathrm{Red})}
\end{equation}
As all three sensors used in BaKIM capture near-infrared and red bands, we can calculate the NDVI for all AOIs.
\vspace{10pt}

\noindent \textbf{NDRE:} The Normalised Difference Red Edge index takes the red edge (RE) band as well as the red band into account and represents the chlorophyll content in leaves~\cite{barnes2000coincident}. In their review of commonly used remote sensing technologies to measure plant water stress, Govender et al. found red edge to be one of the most important bands when investing plant stress~\cite{govender2009review}. The NDRE index is calculated as follows:
\begin{equation}
    \mathrm{NDRE}=\frac{(\mathrm{RE}-\mathrm{Red})}{(\mathrm{RE}+\mathrm{Red})}
\end{equation}
As all three sensors used in BaKIM capture red edge and red bands, we can calculate the NDRE for all AOIs.
\vspace{10pt}

Additionally to the NDVI and NDRE, the thermal band of the \emph{Micasense Altum-PT} sensor theoretically allows us to use further indices that reflect water stress in plants. As the calculation for such indices is often based on additional ground measurements, we still have to see in how far we can implement them in BaKIM~\cite{govender2009review}.

\subsection{Deep Learning for Tree Vitality Assessment}%
A second option for tree vitality assessment from very-high-resolution UAV data is Deep Learning. On the one hand, mistletoe (posing a secondary parasite) can easily be seen in UAV data, and on the other, dead leaves and branches in tree crowns are also visible in UAV data. Therefore, CNNs, as introduced before, can be used to classify mistletoe or the vitality of tree crowns~\cite{miraki2021detection}. The only problem is, again, the need for a high amount of training data. Together with an apprentice of the forestry department, we plan to gather ground truth data on mistletoe visible in our UAV data and train an object detection CNN to classify mistletoe. The rough estimation of tree vitality labelled by our commissioned forester, as described in Sect.~\ref{sec:ground-truth-labelling}, will be used for exploratory tests using image classification CNNs on images of single delineated tree crowns.

%% file: Content/5-Interactive-Web-Application.tex
\section{Interactive Web Application to Visualise Generated Information}%
Everything described so far is useless if Bamberg's arborists and foresters are not enabled to access the information we generated. To create this access, we chose \emph{dash}\footnote{\url{https://dash.plotly.com/}} to create an interactive web application which visualises all information gathered and generated in BaKIM. The main goal of this web application is to be very flexible so that the arborists and foresters can tailor the underlying data to their needs. This is accomplished by individual filter options and different views and plots. For the storage and filtering of the data \emph{geopandas}\footnote{\url{https://geopandas.org/en/stable/}} is used, for plotting \emph{plotly}\footnote{\url{https://plotly.com/python/}} is used, and to visualise the orthomosaics a \emph{mbtileserver}\footnote{\url{https://github.com/consbio/mbtileserver}} is used. We are currently developing the web application prototype with basic functionalities.

%% file: Content/6-Conclusion.tex
\section{Conclusion and Outlook}%
So far, BaKIM reached important milestones on its way to become a helpful tool based on human-centred AI. The feedback of Bamberg's lead arborist and forester shows that the concept and solutions we are developing promise to be helpful for their daily task planning. Furthermore, the creation of a database of UAV data and predictions, as well as the beginning of tracking the changes made to the tree cadastre, will make time series analysis possible in the future. This will be especially important when it comes to adapting to the climate crisis.

Another benefit of BaKIM is that the infrastructure being built in the project can have multiple uses. For example, the soil moisture sensors installed in Bamberg can also give the arborists live information on the city trees' water deficit. A modified view in the web application could be used to inform Bamberg's citizens about the trees in the city. On top of this, community services like tree watering patronages or an interface to report branches that threaten to fall might be possible in the near future due to BaKIM.

While our tree inventory approaches, with accuracies of about 80\%, are not perfect yet, we already plan several changes and improvements which should yield significantly higher accuracies. Nevertheless, even with higher accuracies, BaKIM will by no means be a substitution for the expertise and decision-making competency of the arborists and foresters. Much rather, the higher frequency and detail of information enables them to improve their work and, thereby, the health of trees in and around Bamberg.

%% file: main.bbl
\begin{thebibliography}{10}
\providecommand{\url}[1]{\texttt{#1}}
\providecommand{\urlprefix}{URL }
\providecommand{\doi}[1]{https://doi.org/#1}

\bibitem{agvolution}
{Agvolution}: Soil moisture sensor. \url{https://www.agvolution.com/} (2023)

\bibitem{pleiades}
Airbus: pleiades neo.
  \url{https://www.eoportal.org/satellite-missions/pleiades-neo} (2023)

\bibitem{Ball2022}
Ball, J.G.C., Hickman, S.H.M., Jackson, T.D., Koay, X.J., Hirst, J., Jay, W.,
  Aubry-Kientz, M., Vincent, G., Coomes, D.A.: Accurate tropical forest
  individual tree crown delineation from {RGB} imagery using mask r-{CNN}.
  bioRxiv  (2022). \doi{10.1101/2022.07.10.499480}

\bibitem{barnes2000coincident}
Barnes, E., Clarke, T., Richards, S., Colaizzi, P., Haberland, J., Kostrzewski,
  M., Waller, P., Choi, C., Riley, E., Thompson, T., et~al.: Coincident
  detection of crop water stress, nitrogen status and canopy density using
  ground based multispectral data. In: Proceedings of the Fifth International
  Conference on Precision Agriculture, Bloomington, MN, USA. vol.~1619, p.~6
  (2000)

\bibitem{hess-drought}
Boeing, F., Rakovec, O., Kumar, R., Samaniego, L., Schr\"on, M., Hildebrandt,
  A., Rebmann, C., Thober, S., M\"uller, S., Zacharias, S., Bogena, H.,
  Schneider, K., Kiese, R., Attinger, S., Marx, A.: High-resolution drought
  simulations and comparison to soil moisture observations in germany.
  Hydrology and Earth System Sciences  \textbf{26}(19),  5137--5161 (2022).
  \doi{10.5194/hess-26-5137-2022},
  \url{https://hess.copernicus.org/articles/26/5137/2022/}

\bibitem{chen2017deeplab}
Chen, L.C., Papandreou, G., Kokkinos, I., Murphy, K., Yuille, A.L.: Deeplab:
  Semantic image segmentation with deep convolutional nets, atrous convolution,
  and fully connected crfs. IEEE transactions on pattern analysis and machine
  intelligence  \textbf{40}(4),  834--848 (2017)

\bibitem{chen2017rethinking}
Chen, L.C., Papandreou, G., Schroff, F., Adam, H.: Rethinking atrous
  convolution for semantic image segmentation. arXiv preprint arXiv:1706.05587
  (2017)

\bibitem{chen2018encoder}
Chen, L.C., Zhu, Y., Papandreou, G., Schroff, F., Adam, H.: Encoder-decoder
  with atrous separable convolution for semantic image segmentation. In:
  Proceedings of the European conference on computer vision (ECCV). pp.
  801--818 (2018)

\bibitem{Cohn2021}
Cohn, R., Holm, E.: Unsupervised machine learning via transfer learning and
  k-means clustering to classify materials image data. Integrating Materials
  and Manufacturing Innovation  \textbf{10}(2),  231--244 (2021).
  \doi{10.1007/s40192-021-00205-8}

\bibitem{dji-m3m}
{DJI}: M3m. \url{https://ag.dji.com/de/mavic-3-m} (2023)

\bibitem{Erikson2005}
Erikson, M., Olofsson, K.: Comparison of three individual tree crown detection
  methods. Machine Vision and Applications  \textbf{16}(4),  258--265 (2005).
  \doi{10.1007/s00138-005-0180-y}

\bibitem{eu-commission}
{European Commission}: Easy access rules for unmanned aircraft systems
  (regulations (eu) 2019/947 and 2019/945) - revision from september 2022.
  \url{https://www.easa.europa.eu/en/document-library/easy-access-rules/easy-access-rules-unmanned-aircraft-systems-regulations-eu}
  (2022)

\bibitem{ferreira2020individual}
Ferreira, M.P., de~Almeida, D.R.A., de~Almeida~Papa, D., Minervino, J.B.S.,
  Veras, H.F.P., Formighieri, A., Santos, C.A.N., Ferreira, M.A.D., Figueiredo,
  E.O., Ferreira, E.J.L.: Individual tree detection and species classification
  of amazonian palms using uav images and deep learning. Forest Ecology and
  Management  \textbf{475},  118397 (2020)

\bibitem{afz-waldzustand}
Fischer, J.: Waldzustand 2022. {AFZ} Der Wald  \textbf{{7/2023}}(7),  35--36
  (2023),
  \url{https://www.digitalmagazin.de/marken/afz-derwald/hauptheft/2023-7/waldschutz/035_waldzustand-2022}

\bibitem{Franklin2018}
Franklin, S.E.: Pixel- and object-based multispectral classification of forest
  tree species from small unmanned aerial vehicles. Journal of Unmanned Vehicle
  Systems  \textbf{6}(4),  195--211 (2018). \doi{10.1139/juvs-2017-0022}

\bibitem{Gini2014}
Gini, R., Passoni, D., Pinto, L., Sona, G.: Use of unmanned aerial systems for
  multispectral survey and tree classification: a test in a park area of
  northern italy. European Journal of Remote Sensing  \textbf{47}(1),  251--269
  (2014). \doi{10.5721/eujrs20144716}

\bibitem{govender2009review}
Govender, M., Govender, P., Weiersbye, I., Witkowski, E., Ahmed, F.: Review of
  commonly used remote sensing and ground-based technologies to measure plant
  water stress. Water Sa  \textbf{35}(5) (2009)

\bibitem{Gradisar2023}
Gradi{\v{s}}ar, L., Dolenc, M.: Transfer and unsupervised learning: An
  integrated approach to concrete crack image analysis. Sustainability
  \textbf{15}(4), ~3653 (2023). \doi{10.3390/su15043653}

\bibitem{Hanapi2019}
Hanapi, S.N.H.S., Shukor, S.A.A., Johari, J.: A review on remote sensing-based
  method for tree detection and delineation. {IOP} Conference Series: Materials
  Science and Engineering  \textbf{705}(1),  012024 (2019).
  \doi{10.1088/1757-899x/705/1/012024}

\bibitem{Hao2021}
Hao, Z., Lin, L., Post, C.J., Mikhailova, E.A., Li, M., Chen, Y., Yu, K., Liu,
  J.: Automated tree-crown and height detection in a young forest plantation
  using mask region-based convolutional neural network (mask r-{CNN}). {ISPRS}
  Journal of Photogrammetry and Remote Sensing  \textbf{178},  112--123 (2021).
  \doi{10.1016/j.isprsjprs.2021.06.003}

\bibitem{holschneider1990real}
Holschneider, M., Kronland-Martinet, R., Morlet, J., Tchamitchian, P.: A
  real-time algorithm for signal analysis with the help of the wavelet
  transform. In: Wavelets: Time-Frequency Methods and Phase Space Proceedings
  of the International Conference, Marseille, France, December 14--18, 1987.
  pp. 286--297. Springer (1990)

\bibitem{Huo2020}
Huo, L., Lindberg, E.: Individual tree detection using template matching of
  multiple rasters derived from multispectral airborne laser scanning data.
  International Journal of Remote Sensing  \textbf{41}(24),  9525--9544 (2020).
  \doi{10.1080/01431161.2020.1800127}

\bibitem{huo2023assessing}
Huo, L., Lindberg, E., Bohlin, J., Persson, H.J.: Assessing the detectability
  of european spruce bark beetle green attack in multispectral drone images
  with high spatial-and temporal resolutions. Remote Sensing of Environment
  \textbf{287},  113484 (2023)

\bibitem{ismail2007forest}
Ismail, R., Mutanga, O., Bob, U.: Forest health and vitality: the detection and
  monitoring of pinus patula trees infected by sirex noctilio using digital
  multispectral imagery. Southern Hemisphere Forestry Journal  \textbf{69}(1),
  39--47 (2007)

\bibitem{Ji2022}
Ji, M., Zhong, J., Xue, R., Su, W., Kong, Y., Fei, Y., Ma, J., Wang, Y., Mi,
  L.: Early detection of cervical cancer by fluorescence lifetime imaging
  microscopy combined with unsupervised machine learning. International Journal
  of Molecular Sciences  \textbf{23}(19),  11476 (2022).
  \doi{10.3390/ijms231911476}

\bibitem{kattenborn2021review}
Kattenborn, T., Leitloff, J., Schiefer, F., Hinz, S.: Review on convolutional
  neural networks (cnn) in vegetation remote sensing. ISPRS journal of
  photogrammetry and remote sensing  \textbf{173},  24--49 (2021)

\bibitem{kehr-Folgeschaeden}
Kehr, R.: Possible effects of drought stress on native broadleaved tree species
  - assessment in light of the 2018/19 drought. Jahrbuch der Baumpflege 2020
  pp. 103--107 (2020)

\bibitem{Larsen2011}
Larsen, M., Eriksson, M., Descombes, X., Perrin, G., Brandtberg, T., Gougeon,
  F.A.: Comparison of six individual tree crown detection algorithms evaluated
  under varying forest conditions. International Journal of Remote Sensing
  \textbf{32}(20),  5827--5852 (2011). \doi{10.1080/01431161.2010.507790}

\bibitem{lausch2013forecasting}
Lausch, A., Heurich, M., Gordalla, D., Dobner, H.J., Gwillym-Margianto, S.,
  Salbach, C.: Forecasting potential bark beetle outbreaks based on spruce
  forest vitality using hyperspectral remote-sensing techniques at different
  scales. Forest Ecology and Management  \textbf{308},  76--89 (2013)

\bibitem{Liao2022}
Liao, L., Cao, L., Xie, Y., Luo, J., Wang, G.: Phenotypic traits extraction and
  genetic characteristics assessment of eucalyptus trials based on {UAV}-borne
  {LiDAR} and {RGB} images. Remote Sensing  \textbf{14}(3), ~765 (2022).
  \doi{10.3390/rs14030765}

\bibitem{lobo2020applying}
Lobo~Torres, D., Queiroz~Feitosa, R., Nigri~Happ, P., Elena Cu{\'e} La~Rosa,
  L., Marcato~Junior, J., Martins, J., Ol{\~a}~Bressan, P., Gon{\c{c}}alves,
  W.N., Liesenberg, V.: Applying fully convolutional architectures for semantic
  segmentation of a single tree species in urban environment on high resolution
  uav optical imagery. Sensors  \textbf{20}(2), ~563 (2020)

\bibitem{Lucena2022}
Lucena, F., Breunig, F.M., Kux, H.: The combined use of {UAV}-based {RGB} and
  {DEM} images for the detection and delineation of orange tree crowns with
  mask r-{CNN}: An approach of labeling and unified framework. Future Internet
  \textbf{14}(10), ~275 (2022). \doi{10.3390/fi14100275}

\bibitem{miraki2021detection}
Miraki, M., Sohrabi, H., Fatehi, P., Kneubuehler, M.: Detection of mistletoe
  infected trees using uav high spatial resolution images. Journal of Plant
  Diseases and Protection  \textbf{128},  1679--1689 (2021)

\bibitem{monnet2010}
Monnet, J.M., Mermin, E., Chanussot, J., Berger, F.: Tree top detection using
  local maxima filtering: a parameter sensitivity analysis. In: 10th
  International Conference on LiDAR Applications for Assessing Forest
  Ecosystems (Silvilaser 2010) (2010)

\bibitem{morales2018automatic}
Morales, G., Kemper, G., Sevillano, G., Arteaga, D., Ortega, I., Telles, J.:
  Automatic segmentation of mauritia flexuosa in unmanned aerial vehicle
  ({UAV}) imagery using deep learning. Forests  \textbf{9}(12), ~736 (2018)

\bibitem{moran1994estimating}
Moran, M., Clarke, T., Inoue, Y., Vidal, A.: Estimating crop water deficit
  using the relation between surface-air temperature and spectral vegetation
  index. Remote sensing of environment  \textbf{49}(3),  246--263 (1994)

\bibitem{Pitkaenen2001}
Pitkänen, J.: Individual tree detection in digital aerial images by combining
  locally adaptive binarization and local maxima methods. Canadian Journal of
  Forest Research  \textbf{31}(5),  832--844 (2001). \doi{10.1139/x01-013}

\bibitem{trinity}
{Quantum Systems}: Trinity {F90+}.
  \url{https://quantum-systems.com/trinity-f90/} (2023)

\bibitem{ipcc2022}
Rama, H.O., Roberts, D., Tignor, M., Poloczanska, E., Mintenbeck, K., Alegría,
  A., Craig, M., Langsdorf, S., Löschke, S., Möller, V., Okem, A., Rama, B.,
  Ayanlade, S.: Climate Change 2022: Impacts, Adaptation and Vulnerability
  Working Group II Contribution to the Sixth Assessment Report of the
  Intergovernmental Panel on Climate Change. Cambridge University Press (2022).
  \doi{10.1017/9781009325844.}

\bibitem{Raum-Auswirkungen}
Raum, S., Collins, M.C., Urquhart, J., Potter, C., Pauleit, S., egerer, M.: Die
  vielf\"altigen auswirkungen von baumsch\"adlingen und krankheitserregern im
  urbanen raum. Pro{BAUM}  \textbf{{1/2023}},  103--107 (2023)

\bibitem{ronneberger2015u}
Ronneberger, O., Fischer, P., Brox, T.: U-net: Convolutional networks for
  biomedical image segmentation. In: Medical Image Computing and
  Computer-Assisted Intervention--MICCAI 2015: 18th International Conference,
  Munich, Germany, October 5-9, 2015, Proceedings, Part III 18. pp. 234--241.
  Springer (2015)

\bibitem{rouse1974monitoring}
Rouse~Jr, J.W., Haas, R.H., Deering, D., Schell, J., Harlan, J.C.: Monitoring
  the vernal advancement and retrogradation (green wave effect) of natural
  vegetation (1974)

\bibitem{samek2017explainable}
Samek, W., Wiegand, T., M{\"u}ller, K.R.: Explainable artificial intelligence:
  Understanding, visualizing and interpreting deep learning models. arXiv
  preprint arXiv:1708.08296  (2017)

\bibitem{schiefer-fortress}
Schiefer, F., Frey, J., Kattenborn, T.: Fortress: Forest tree species
  segmentation in very-high resolution uav-based orthomosaics (2022).
  \doi{10.35097/538}

\bibitem{schiefer2020mapping}
Schiefer, F., Kattenborn, T., Frick, A., Frey, J., Schall, P., Koch, B.,
  Schmidtlein, S.: Mapping forest tree species in high resolution uav-based
  rgb-imagery by means of convolutional neural networks. ISPRS Journal of
  Photogrammetry and Remote Sensing  \textbf{170},  205--215 (2020)

\bibitem{Schaefer2016}
Schäfer, E., Heiskanen, J., Heikinheimo, V., Pellikka, P.: Mapping tree
  species diversity of a tropical montane forest by unsupervised clustering of
  airborne imaging spectroscopy data. Ecological Indicators  \textbf{64},
  49--58 (2016). \doi{10.1016/j.ecolind.2015.12.026}

\bibitem{senoo1988assessment}
SENOO, T., HONJYO, T.: Assessment of tree stress by airborne multi-spectral
  scanning data. Journal of the Japanese Forestry Society  \textbf{70}(2),
  45--56 (1988)

\bibitem{waefler2020explainability}
Waefler, T., Schmid, U.: Explainability is not enough: {Requirements} for
  human-{AI}-partnership in complex socio-technical systems. In: Proceedings of
  the 2nd European Conference on the Impact of Artificial Intelligence and
  Robotics (ECIAIR 2020). pp. 185--194. ACPIL (2020)

\bibitem{wang2004individual}
Wang, L., Gong, P., Biging, G.S.: Individual tree-crown delineation and treetop
  detection in high-spatial-resolution aerial imagery. Photogrammetric
  Engineering \& Remote Sensing  \textbf{70}(3),  351--357 (2004)

\bibitem{Weinstein2020a}
Weinstein, B.G., Marconi, S., Aubry-Kientz, M., Vincent, G., Senyondo, H.,
  White, E.P.: Deepforest: A python package for rgb deep learning tree crown
  delineationdeepforest: A python package for rgb deep learning tree crown
  delineation. Methods in Ecology and Evolution  \textbf{11}(12),  1743--1751
  (2020). \doi{10.1111/2041-210x.13472}

\bibitem{Yang2022}
Yang, M., Mou, Y., Liu, S., Meng, Y., Liu, Z., Li, P., Xiang, W., Zhou, X.,
  Peng, C.: Detecting and mapping tree crowns based on convolutional neural
  network and google earth images. International Journal of Applied Earth
  Observation and Geoinformation  \textbf{108},  102764 (2022).
  \doi{10.1016/j.jag.2022.102764}

\bibitem{Yu2022}
Yu, K., Hao, Z., Post, C.J., Mikhailova, E.A., Lin, L., Zhao, G., Tian, S.,
  Liu, J.: Comparison of classical methods and mask r-{CNN} for automatic tree
  detection and mapping using {UAV} imagery. Remote Sensing  \textbf{14}(2),
  ~295 (2022). \doi{10.3390/rs14020295}

\end{thebibliography}
